\documentclass{article}
\usepackage{spconf,amsmath,graphicx, soul, color}
\usepackage{enumitem}

\title{HRFNet: High-Resolution Forgery Network for Localizing Satellite Image Manipulation}
\name{Fahim Faisal Niloy$^{\dagger}$ \qquad Kishor Kumar Bhaumik$^{\ddagger}$ \qquad Simon S. Woo$^{\ddagger}$}
  
\address{$^{\dagger}$University of California, Riverside \\ $^{\ddagger}$Sungkyunkwan University, South Korea}
      
\begin{document}
%\ninept
%
\maketitle
%
% Recently, with the development of vision transformers, object representations are becoming more effective for various downstream vision tasks. Deepfake videos are mostly generated in a frame-by-frame manner, which leaves visible object-level inconsistencies in both temporal and spatial dimensions. Object representations can also be an important clue for the detection of deepfake videos. 

\begin{abstract}
Existing high-resolution satellite image forgery localization methods rely on patch-based or downsampling-based training. Both of these training methods have major drawbacks, such as inaccurate boundaries between pristine and forged regions, the generation of unwanted artifacts, etc. To tackle the aforementioned challenges, inspired by the high-resolution image segmentation literature, we propose a novel model called \textit{HRFNet} to enable satellite image forgery localization effectively. Specifically, equipped with shallow and deep branches, our model can successfully integrate RGB and resampling features in both global and local manners to localize forgery more accurately. We perform various experiments to demonstrate that our method achieves the best performance, while the memory requirement and processing speed are not compromised compared to existing methods.

\end{abstract}
\section{Introduction}
\label{sec:intro}
The advent of satellites equipped with advanced imaging technology has enabled a wide range of applications that utilize satellite imagery, such as agricultural crop classification \cite{gao2019new}, wildlife monitoring, etc. Concurrently, the proliferation of image and video editing software, along with advanced AI techniques such as deepfakes \cite{add, tariq2022real}, has led to the creation of fake images. Consequently, image forgery has emerged as a significant socio-technical concern. Satellite images are also susceptible to manipulation using these tools, highlighting the need for automated methods to detect tampered areas. However, accurately identifying tampered areas with different types of forgery (including splicing, copy-move, removal, etc.) remains highly challenging.

Typically, satellite images are high-resolution images with large spatial sizes. On the other hand, existing general-purpose image forgery localization methods are designed for lower-resolution images. Therefore, using these methods directly on high-resolution satellite images is not feasible due to resource constraints. To address this issue, several approaches \cite{horvath2020manipulation, horvath2021manipulation, bartusiak2019splicing, horvath2019anomaly} specifically targeting satellite image forgery localization have been proposed. These approaches often involve patch extraction from the satellite image or downsampling the high-resolution image before inputting it to the forgery localization network for segmentation. 

%The global branch helps the local branch with the global context information and the local branch helps the global branch with more low-level fine-grained information.

However, such approaches have their limitations. Chen et al. \cite{chen2019collaborative} have shown that patch-based or downsampling-based training for the segmentation of high-resolution images has major drawbacks. Specifically, patch-based training lacks spatial contexts and neighborhood dependency information, making it difficult to distinguish between different classes. On the other hand, downsampling-based training suffers from ``jiggling" artifacts and inaccurate boundaries due to the missing details from downsampling. To further tackle these challenges and facilitate high-resolution image segmentation on satellite images, a few approaches have been proposed. For example, GLNet \cite{chen2019collaborative} uses a downsampled version of the full-resolution input image in the global branch and extracts patches from the same input image into the local branch, with the aim that the two branches will collaborate with each other. MBNet \cite{shan2022mbnet} improves upon GLNet by taking co-centered patches of various sizes, instead of only one resolution. In addition, ISDNet \cite{guo2022isdnet} takes the full-resolution image as input but uses a lightweight network to extract features. To improve the low-level fine-grained features extracted by the lightweight network, a deep network is also used to extract features from the downsampled full-resolution image in a parallel branch.

%After that, the features from the deep network are fused with the features of shallow network and then sent to the decoder for segmentation. 

\begin{figure*}[!t]
\centering
\includegraphics[width=0.8\textwidth]{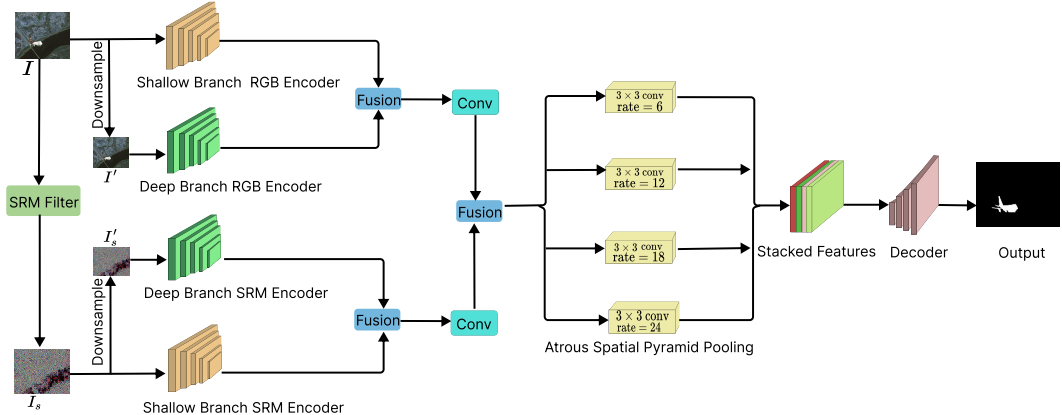}  %{hfnet.png}
% \vspace{-15pt}
\caption{Overview of proposed HRFNet: The shallow branch consisting of a lightweight model takes the full resolution satellite image as input. In contrast, the deep branch consisting of a deep network takes the downsampled satellite image as input. Both RGB and resampling features are extracted and fused to output the final prediction mask.}
\label{fig:hfnet}
\end{figure*}

However, methods like ISDNet and MBNet are specifically tailored for the segmentation of high-resolution images and do not readily translate for usage in the case of satellite image forgery localization. This is because forgery localization, although a segmentation task, requires additional features to accurately pinpoint the forged region within the image. Moreover, no attempt has been made yet to significantly deviate from the traditional patch-based or downsampling-based training approaches in this field. As a result of all the above factors and challenges, existing satellite forgery localization methods still suffer from drawbacks such as inaccurate boundaries and the generation of artifacts.

Our work aims to overcome such drawbacks by leveraging training strategy from high-resolution image segmentation literature and tailoring and translating it to fit the satellite image manipulation localization task. Specifically, inspired by the shallow and deep integration of ISDNet \cite{guo2022isdnet}, we propose a novel model called ``HRFNet", which successfully integrates resampling and RGB features from both the shallow branch and deep branch to effectively perform satellite image manipulation localization. The RGB branches assist in distinguishing tampered regions from authentic regions by capturing visual inconsistencies at tampered boundaries, such as the contrast effect between tampered regions and authentic regions, etc. On the other hand, SRM branches analyze the local noise features in an image, such as resampling artifacts, etc.

Both the RGB and SRM branches are further divided into shallow and deep branches. The shallow branch enables the extraction of features in a global manner with a wide field of view, whereas the deep branch extracts more detailed and hierarchical features. By fusing features from both branches, our model successfully captures image manipulation traces and becomes effective in localizing high-resolution satellite image manipulation in an efficient manner. It should also be noted that we only adopt the training strategy of ISDNet, which is the local-global collaboration strategy. However, the underlying architecture of HRFNet is fundamentally different compared to ISDNet. In summary, our main contributions are:

% The above methodes are usually for segmentation of vision tasks. We specially modify the architecture to tailor it for forgery localization task by adding SRM branch and fusion.
\begin{itemize}[leftmargin=*,topsep=0pt]
\setlength{\itemsep}{-2pt}
\item We propose a novel satellite image manipulation localization model named \textit{HRFNet} that is equipped with RGB and SRM branches of shallow and deep nature for successfully integrating both global contextual information and fine-grained local information to capture forgery traces from high-resolution satellite images effectively. 
\item Our approach deviates from the conventional patch-based or downsampling-based methods, specifically in the case of satellite image manipulation localization task, aiming to tackle the drawbacks inherent to the previous approaches.
\item We perform several experiments on satellite manipulation dataset to show that our method achieves the highest performance compared to existing methods, while the memory footprint and processing speed are not compromised.
\end{itemize}

\noindent \textbf{Related Works: }
Extensive research has been conducted on generic image forgery localization, which is not specific to satellite images. ManTra-Net \cite{wu2019mantra} is a prominent forgery localization model that uses a VGG-based feature extractor and an LSTM-based detection module. It is trained to detect various types of image manipulation traces. Moreover, CFLNet \cite{niloy2023cfl} combines contrastive learning with cross-entropy loss for more generalizable forgery localization. However, these methods are designed for lower-resolution images. Therefore, downsampling or patch extraction is necessary when applied to manipulated satellite images, which reduces performance.

On the other hand, several research works have been proposed which are specifically focused on satellite image manipulation localization. One approach by \cite{horvath2020manipulation} extracts patches from full-resolution satellite images and uses a deep belief network (DBN) to reconstruct them. The model detects and localizes forgery based on the reconstruction error, which becomes significant for manipulated images. Authors in \cite{bartusiak2019splicing} employ a conditional GAN, where the generator aims to produce the ground truth mask and the discriminator classifies forgery using real image-mask pairs and synthetic pairs. In \cite{horvath2019anomaly}, authors use an auto-encoder to reconstruct patches from satellite images and apply an SVDD-based one-class classifier to detect and localize forgery.

% It should be noted that, image manipulation detection is the task of classifying whether an image is forged or not. Whereas, image manipulation localization is a segmentation task, which additionally localizes the manipulated region.

\section{Method}
\label{sec:method}

In Fig. \ref{fig:hfnet}, we present the overall diagram of our proposed HRFNet, where the inputs are the full resolution RGB image $I$ and its downsampled version $I'$. As $I$ represents the full resolution image, using deep networks is impractical due to memory constraints. Hence, we utilize a lightweight network in the shallow RGB branch to extract features from $I$. However, a shallow network performs poorly in capturing long-range and high-level semantic cues. To complement the shallow branch features, we introduce a deep network in the deep RGB branch, which takes $I'$ as input. The deep network effectively extracts high-level features from downsampled images. However, the deep network loses spatial information due to the downsampled input image, which is better captured by the shallow network using the full resolution image input.

Next, we incorporate a fusion module to merge features from the shallow and deep RGB branches, aiming to enhance segmentation performance through their complementary nature. While sophisticated fusion methods have been proposed in the literature \cite{guo2022isdnet, fu2019dual}, it is important to note that our paper primarily focuses on introducing a novel training strategy for satellite image manipulation localization. Hence, we adopt a simpler fusion mechanism that involves concatenating the features.

%It should be noted that, sophisticated fusion methods exist \cite{guo2022isdnet, fu2019dual}, However, our paper focuses on the new training strategy for satellite image manipulation localization. Hence, we opt for simple fusing mehcanism involving concatenating of the features.

% Further research can explore advanced fusion mechanisms to improve performance further.

Additionally, we utilize SRM filters \cite{fridrich2012rich} on $I$ to generate the output $I_{s}$. SRM filters are high pass filters that capture resampling features and enhance the high-frequency information of the input image, which is beneficial for forgery localization. $I_{s}$ is then downsampled to obtain $I_{s}'$. Similarly, $I_{s}$ and $I_{s}'$ are passed through the shallow and deep SRM branches respectively. Subsequently, feature concatenation is again employed to fuse the features from both SRM branches.

Following the fusion module, we apply a conv-relu-conv layer to the final features from the RGB and SRM branches, followed by another fusion module that concatenates features, combining the information from both branches. An ASPP module \cite{chen2017rethinking} is then utilized on the resulting fused feature maps to extract multi-scale information. It has been reported in \cite{zhou2018learning} that global context aids in gathering more clues for manipulation detection. The ASPP module contributes to this by extracting information at different scales, thereby making global context and fine-grained pixel-level context available. The output of the ASPP module is passed to the decoder, where we employ a DeepLabv3+ style segmentation head to generate the final segmentation mask.

\section{Experiments}
% In this section we describe our dataset and implementation details, and explain our experimental results.

\begin{table}[t]

\centering
\caption{AUC Scores (in \%).}

\begin{tabular}{ll}
\hline
Methods                       & AUC (\%)            \\ \hline \hline
MantraNet (CVPR'19)           & 77.31          \\
DBN (CVPRW'20)                & 63.80          \\
Vision Transformer (CVPRW'21) & 73.26          \\
CFLNet (WACV'23)              & 83.52          \\
HRFNet \textit{\textbf{(ours)}}                          & \textbf{87.36} \\ \hline
\end{tabular}
\label{table:auc}
\end{table}

\noindent \textbf{Dataset:} There is a scarcity of benchmark high-resolution satellite image manipulation datasets. Hence, we take the benchmark satellite dataset Deepglobe \cite{demir2018deepglobe} and follow the process from \cite{horvath2020manipulation} to manipulate the satellite images. Each final manipulated image in the dataset has a resolution of $1,000 \times 1,000$, as described in \cite{horvath2020manipulation}. 

\noindent \textbf{Baseline Models:} We compare our method with methods that are specifically designed for satellite image manipulation localization - Vision transformer with post processing \cite{horvath2021manipulation} and DBN \cite{horvath2020manipulation}. We also compare with the prominent image forgery localization method - MantraNet \cite{wu2019mantra} and the current SOTA image forgery localization model CFLNet \cite{niloy2023cfl}. For MantraNet and CFLNet, we resize input images to the size specified by their respective papers for comparisons.

\noindent \textbf{Implementation Details:} We use ResNet-18 as a network backbone for both deep RGB and deep SRM branches. MobileNet-v3 is used for the shallow RGB and shallow SRM branches. For the deep networks, we downsample the input image to $224 \times 224$ pixel-size. We train HRFNet with Adam optimizer with a learning rate of $1e-3$. We reduce the learning rate by 20\% after every 20 epochs and train for 100 epochs. Cross-entropy loss is weighted to provide the tampered class ten times more weight \cite{niloy2023cfl}. We set the batch size to 4 and train the model on NVIDIA Tesla K80 GPU.

\noindent \textbf{Results:} We report the AUC scores (in \%) of our method and the baseline models in Table \ref{table:auc}. As shown, our method achieves the highest performance compared to the other methods, yielding a $3.84\%$ AUC improvement over the current state-of-the-art methods for generic image manipulation localization. One reason behind the poor performance of Vision Transformer and DBN is that both the methods generate heatmap and then threshold on the heatmap to generate the output mask instead of using any learnable decoder.

To compare the resource requirement for HRFNet, we compute the memory and processing time (in frames per second or FPS) of our model and the best performing baseline models of Table \ref{table:auc}. The objective of this experiment is to demonstrate that HRFNet achieves the best result without requiring higher resource requirements compared to the best baseline models. To measure the GPU memory usage of a model, we use the command line tool “gpustat”, with the minibatch size
of 1, and avoid calculating any gradients as instructed in \cite{chen2019collaborative}.

\begin{table}[t!]
\caption{Memory requirement in MB and processing speeds in FPS. Our method does not require extra resources compared to baseline models while achieving the best performance.}

\centering
\begin{tabular}{ccc}
\hline 
Methods      & Memory (MB) & FPS   \\ \hline \hline
MantraNet    & 682         & 5.18  \\
CFLNet       & 1,740        & 43.12 \\
HRFNet \textit{\textbf{(ours)}} & 1,402        & 33.68 \\ \hline
\end{tabular}

\label{table:resource}
% \vspace{-10pt}
\end{table}

Table \ref{table:resource} shows the memory performance results. Our model achieves comparable results with the best baseline models that do not take the full resolution image as an input. Here, CFLNet requires the most amount of memory. Although we use a two parallel architecture similar to CFLNet, the network in our two deep branches is ResNet-18, whereas CFLNet uses two ResNet-50 based encoders. ResNet-50 has more than twice the number of parameters than ResNet-18. Hence, the overall memory requirement of HRFNet is less than CFLNet. Although MantraNet requires the least amount of memory, the AUC score of MantraNet is also the least among the three models. HRFNet achieves a balanced spot regarding the memory requirement, while the AUC score is the highest. For low-powered mobile devices, processing speed is also a major factor. We have also measured the processing time in FPS for the three models in Table \ref{table:resource}. In fact, CFLNet achieves the best processing speed due to its simpler encoder-decoder type architecture, while HRFNet performs better than MantraNet. Here again, it is demonstrated that our model achieves a balanced spot and does not need extra resources compared to best performing baseline models.
Based on table~\ref{table:resource}, we can easily conclude that the resource requirement for HRFNet is very much comparable to the baseline models, while it achieves the best localization performance.

% \textbf{Ablation Study:} 
% Here we show individual performance of our deep branch and shallow branch. Specifically, we train using only the deep branch and shallow branch individually and show the AUC results on Table \ref{table:ablation}.

% \begin{table}[h!]
% \centering
% \begin{tabular}{ll}
% \hline
% Methods        & AUC   \\ \hline
% Shallow Branch & 77.31 \\
% Deep Branch    & 63.80 \\
% HFNet          & 87.36 \\ \hline
% \end{tabular}
% \caption{Ablation Study}
% \label{table:ablation}
% \end{table}

\begin{figure}[t!]
\setlength\tabcolsep{1pt}%%
\centering
\begin{tabular}{ccccc}
\includegraphics[width=0.8in]{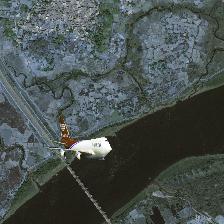} &
 \includegraphics[width=0.8in]{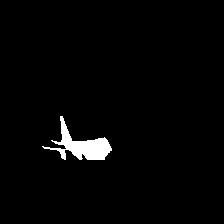} &
 \includegraphics[width=0.8in]{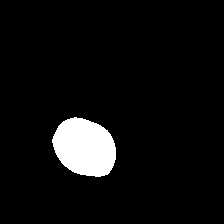} &
 \includegraphics[width=0.8in]{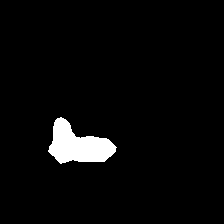} \\
 \includegraphics[width=0.8in]{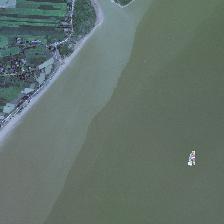} &
 \includegraphics[width=0.8in]{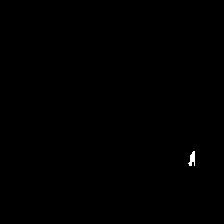} &
 \includegraphics[width=0.8in]{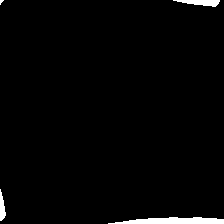} &
 \includegraphics[width=0.8in]{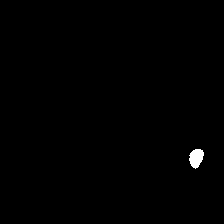} \\
 \includegraphics[width=0.8in]{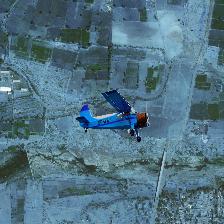} &
 \includegraphics[width=0.8in]{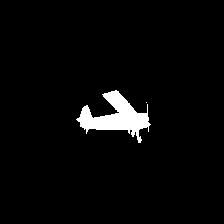} &
 \includegraphics[width=0.8in]{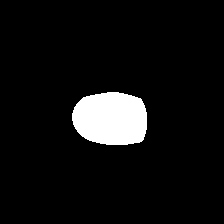} &
 \includegraphics[width=0.8in]{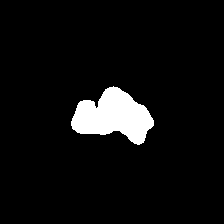}  \\
 \includegraphics[width=0.8in]{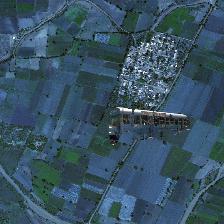} &
 \includegraphics[width=0.8in]{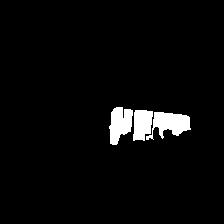} &
 \includegraphics[width=0.8in]{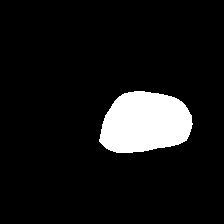} &
 \includegraphics[width=0.8in]{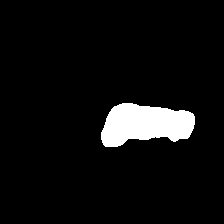} \\
 \textbf{\footnotesize{Input}} &
 \textbf{\footnotesize{GT Mask}} &
 \textbf{\footnotesize{CFLNet}} &
 \textbf{\footnotesize{Ours}} \\
\end{tabular}
\caption{Comparison of the predicted mask with the best performing baseline model - CFLNet. It is evident that the prediction of HRFNet is closer to the ground truth mask compared to CFLNet.}
\label{fig:segmentation}
\end{figure}

\noindent \textbf{Visualization:} In Fig. \ref{fig:segmentation}, we present the sample visualizations of the predicted masks generated by HRFNet and compare them to those of the best performing baseline model in our experiment - CFLNet. It is evident that the predicted masks of CFLNet have inaccurate boundaries compared to HRFNet due to downscaling the input image. This is in accordance with \cite{chen2019collaborative}, where it has been claimed that downsampling high-resolution images results in inaccurate boundaries and unwanted artifacts in the predicted masks. Additionally, when the forged object's shape is very small, CFLNet struggles to localize it, which is expected because downsampling discards important spatial information. On the other hand, the shallow branch of HRFNet extracts features that have accurate spatial information, and the deep branch extracts high-level semantic cues. Complementing both features yields more accurate localization results for the forged objects using our method.

\section{Conclusion}
In this paper, we propose a novel satellite image manipulation localization model, \textit{HRFNet}. Deviating from conventional training strategy that involves patch extraction or downsampling, our method introduces a new training strategy specifically targeted for the task of high-resolution satellite image manipulation localization. Our method exploits shallow and deep integration to successfully capture both global context and high level semantic clues, as well as noise features in order to trace and localize forgery. We have compared our method with existing satellite image localization methods, as well as SOTA generic image manipulation localization methods and demonstrated that our method achieves the best performance while the memory requirement and processing time remain similar to existing methods. We hope our work will inspire more research in this area for effectively localizing satellite image manipulation.

\section*{Acknowledgments}
This work was partly supported by Institute for Information \& communication Technology Planning \& evaluation (IITP) grants funded by the Korean government MSIT: (No. 2022-0-01199, Graduate School of Convergence Security at Sungkyunkwan University), (No. 2022-0-01045, Self-directed Multi-Modal Intelligence for solving unknown, open domain problems), (No. 2022-0-00688, AI Platform to Fully Adapt and Reflect Privacy-Policy Changes), (No. 2021-0-02068, Artificial Intelligence Innovation Hub), (No. 2019-0-00421, AI Graduate School Support Program at Sungkyunkwan University), and (No. RS-2023-00230337, Advanced and Proactive AI Platform Research and Development Against Malicious deepfakes).

\bibliographystyle{IEEEbib}
\bibliography{strings,refs}

\end{document}